\documentclass[a4paper,10pt]{article}
\usepackage[latin1]{inputenc}
\usepackage[english,francais]{babel}
\usepackage{graphics}
\usepackage{subfigure}
\usepackage{pstcol}
\usepackage{amsfonts}
\usepackage{array}
\usepackage{vmargin}

\setmarginsrb{2.1 cm}{2.7 cm}{2.1 cm}{1.7 cm}%
{0 cm}{0 mm}{1.7 cm}{1 cm}

\title{Parametric stiffness analysis of the Orthoglide}

\author{F\'elix Majou$^{1,2}$, Cl\'ement Gosselin$^2$, Philippe Wenger$^1$ and Damien Chablat$^1$ \\
{~}\\ 
{\it$^1$ IRCCyN}
\thanks{Institut de Recherches en Communications et Cybern\'etique de Nantes: UMR CNRS 6597, \'Ecole Centrale de Nantes, \'Ecole des Mines de Nantes, Universit\'e de Nantes}
{\it , 1 rue de la Noë, 44321 Nantes, France}\\
{\it $^2$  D\'epartement de G\'enie M\'ecanique, Universit\'e}
{\it Laval, Qu\'ebec, Canada, G1K 7P4}}

\begin{document}


\maketitle


{\noindent\bf{Abstract}}\\

This paper presents a parametric  stiffness  analysis of the
Orthoglide, a 3-DOF translational Parallel Kinematic Machine.
First, a compliant modeling of the Orthoglide is conducted based
on an existing method. Then stiffness matrix is symbolically
computed. This allows one to easily study the influence of the
geometric design parameters on the matrix elements. Critical links
are displayed. Cutting forces are then modeled so that static
displacements of the Orthoglide tool during slot milling are
symbolically computed. Influence of the geometric design
parameters on the static displacements is checked as well. Other
machining operations can be modeled. This parametric stiffness
 analysis can be applied to any parallel manipulator for which stiffness
  is a critical issue.\\


\section{Introduction}

Parallel manipulators are claimed to offer good stiffness and
accuracy properties, as well as good dynamic performances. This
makes them attractive for innovative machine-tool structures for
high speed machining \cite{Tlusty99}, \cite{Wenger99},
\cite{Majou01}. When a parallel manipulator is intended to become
a Parallel Kinematic Machine (PKM), stiffness becomes a very
important issue in its design \cite{Pritschow97},
\cite{Company02}, \cite{Brogardh02}. In this paper is presented a
parametric stiffness analysis of the Orthoglide, a 3-axis
translational PKM prototype developped at IRCCyN \cite{Wenger00}.

One of the first stiffness analysis methods for parallel
mechanisms is based on a kinetostatic modeling \cite{Gosselin90}.
It proposes to map the stiffness of parallel mechanisms by taking
into account the compliance of the actuated joints. This method is
not appropriate for PKM whose legs, unlike hexapods, are subject
to bending \cite{Kong}. In \cite{Huang02}, a stiffness estimation of a
tripod-based PKM is proposed that solves this problem. However the
stiffness model presented is not general enough. This gap is
filled in \cite{Gosselin02}. The method is based on a
flexible-link lumped parameter model that replaces the compliance
of the links by localized virtual compliant joints and rigid
links. The approach has two differences from that presented in
\cite{Huang02} (i), the way the link compliances are modeled (ii),
the equations allowing the computation of the stiffness model,
that are more general.

This method is applied to the Orthoglide for a parametric
stiffness analysis. The stiffness matrix elements are symbolically
computed. This allows an easy analysis of the influence of the
Orthoglide critical design parameters. No numerical computations
are conducted until graphical results are generated. First we
present the Orthoglide, then the compliant model. The results
showing the influence of the parameters are presented next.

\section{Compliant modeling of the Orthoglide}
\label{Ortho}
\subsection{Kinematic architecture of the Orthoglide}
\begin{figure}[!ht]
\begin{centering}

        {\resizebox{!}{4cm}
        {\includegraphics{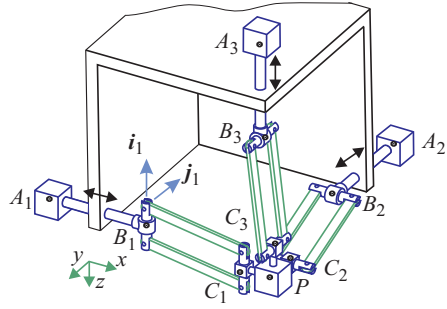}}}

\caption{The Orthoglide kinematic architecture}

\label{ortho_cine}
\end{centering}
\end{figure}

The Orthoglide is a translational 3-axis PKM prototype designed for machining applications.  The mobile platform is connected to three orthogonal linear drives through three identical $RP_aR$ serial chains (where $R$ stands for Revolute and $P_a$ for Parallelogram) (Fig.\ref{ortho_cine}), and it moves in the Cartesian workspace while maintaining a fixed orientation. The Orthoglide is optimized for a prescribed workspace with prescribed kinetostatic performances. Its kinematic analysis, design and optimization are fully described in \cite{Chablat03}.

\subsection{Parameters for compliant modeling}
\begin{figure}[ht!]
\begin{center}
\resizebox{!}{3.5cm}{\includegraphics{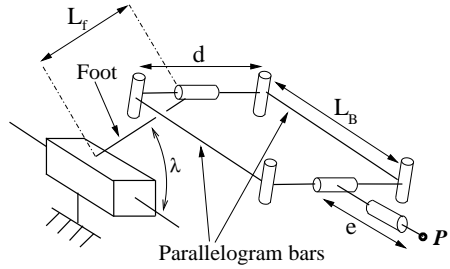}} 
\caption{Leg geometrical parameters}

\label{leg_parameters}
\end{center}
\end{figure}

The parameters used for the compliant  modeling of the Orthoglide are presented in Figure \ref{leg_parameters} and in Table \ref{parameters}. These parameters correspond to a ``beam-like'' modeling of the Orthoglide legs. The foot has been designed to prevent each parallelogram from colliding with the corresponding linear motion guide. \\

\begin{table}[ht!]
\begin{center}

\small{\begin{tabular}[t]{|c|p{5cm}|}

\hline {Parameter} & {Description} \\ \hline

$L_f$ & Foot length, see Fig.\ref{leg_parameters} \\

$h_f$, $b_f$ &  Foot section sides\\

$I_{f_1}=b_f.h_f^3/12$ &  Foot section moment of inertia 1\\

$I_{f_2}=h_f.b_f^3/12$  & Foot section moment of inertia 2\\

$d$  &  Distance between parallelogram bars, see Fig.\ref{leg_parameters} \\

$L_B$ & Parallelogram bar length, see Fig.\ref{leg_parameters} \\

$S_B$ & Parellelogram bar cross-section area \\ \hline

\end{tabular}}
\caption{Geometric parameters} \label{parameters}
\end{center}
\end{table}

\subsection{Compliant modeling with flexible links}
\label{compliance}

\begin{figure}[ht!]
\begin{centering}
\begin{tabular}{c c c}

{\subfigure[Flexible beam]
        {\resizebox{!}{2.5cm}
        {\includegraphics{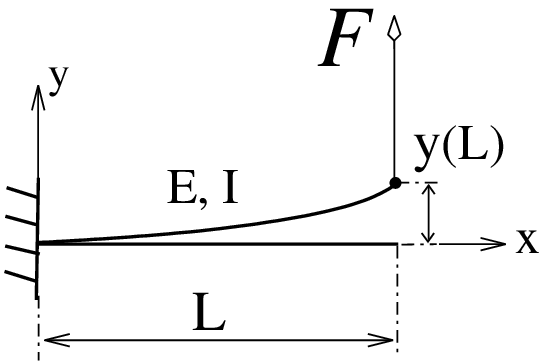}}}}

{\subfigure[Virtual rigid beam]
        {\resizebox{!}{2.3cm}
        {\includegraphics{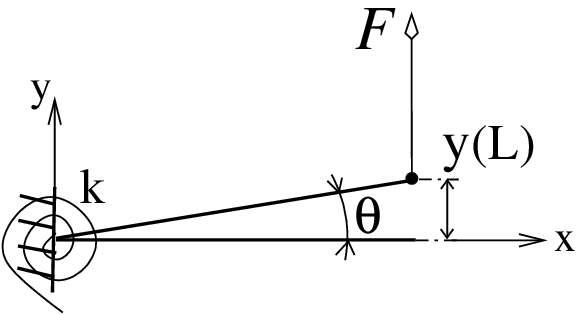}}}}

\end{tabular}
\caption{General model for flexible link}
\label{poutre_flex}
\end{centering}
\end{figure}

In the lumped model described in  \cite{Zhang00}, the leg links
are considered as flexible beams and are  replaced by rigid beams
mounted on revolute joints plus torsional springs located at the
joints (Fig. \ref{poutre_flex}).

From deriving the relationship between the force $F$ and the
deformation $y(x)$, the local torsional stiffness $k$ can be
computed:
\begin{eqnarray}
{EIy''(x)} & {=} & {F(L-x)} \nonumber \\ \nonumber {~} & {\vdots}
& {~} \nonumber \\ \nonumber {EIy(L)} & {=} & {FL^3/3} \nonumber
\\
{\rightarrow \theta \simeq y(L)/L} & {=} & {FL^2/3EI}  \nonumber
\\ {k}  &  {=}  & {FL/\theta}   \nonumber \\  \nonumber
{\rightarrow k}  &  {=}  & {3EI/L}   \nonumber
\end{eqnarray}

\begin{figure}[ht!]
    \begin{center}
          \resizebox{!}{3cm}{\includegraphics{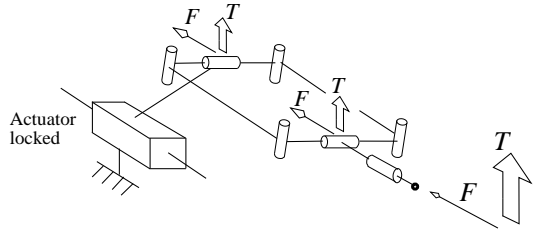}}
           \caption{Forces transmitted in  a leg}
           \label{distri_efforts}
    \end{center}
\end{figure}

If the Orthoglide leg actuator is locked, then one leg can
withstand one force $F$ and one torque $T$
(Fig.~\ref{distri_efforts}), that are transmitted along the
parallelogram bars and the foot. For a compliant modeling that
uses virtual compliant joints, it is important to understand how
external forces are transmitted, and what their effect on the leg
links is.

Three  virtual compliant joints are modeled along the Orthoglide
leg. They are described in Table \ref{tab_flex_first}. The
determination of the virtual joint stiffnesses is not detailed
here to save space. Since the actuator is assumed to be much
stiffer than the virtual compliant joints, it is not included in
foot stiffnesses. The leg links compliances modeled in Table
\ref{tab_flex_first} were selected beforehand as the most
significant. The joint compliances are not taken into account in
our model.

\begin{table}[ht!]
\begin{center}
   \vspace{2mm}
  \small{ \begin{tabular}{|b{3.5cm}|c|} \hline

{\it Stiffness and description} & {\it Figure} \\  \hline

{$k_{act}$: translational stiffness of the prismatic actuator
\newline} &
{\scalebox{.4}{\includegraphics{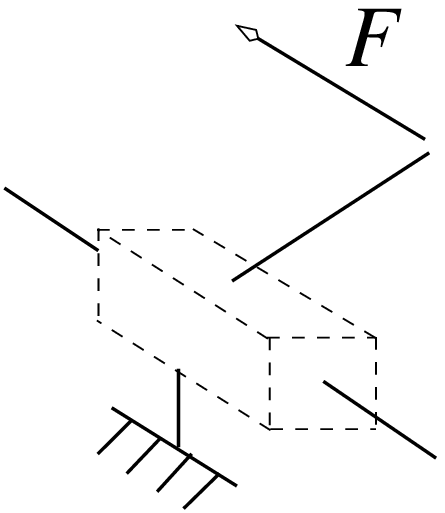}}} \\
\hline

{$2EI_{f_2}/L_f$: foot bending due to torque $T$ \newline {~}
\newline} &
{\scalebox{.4}{\includegraphics{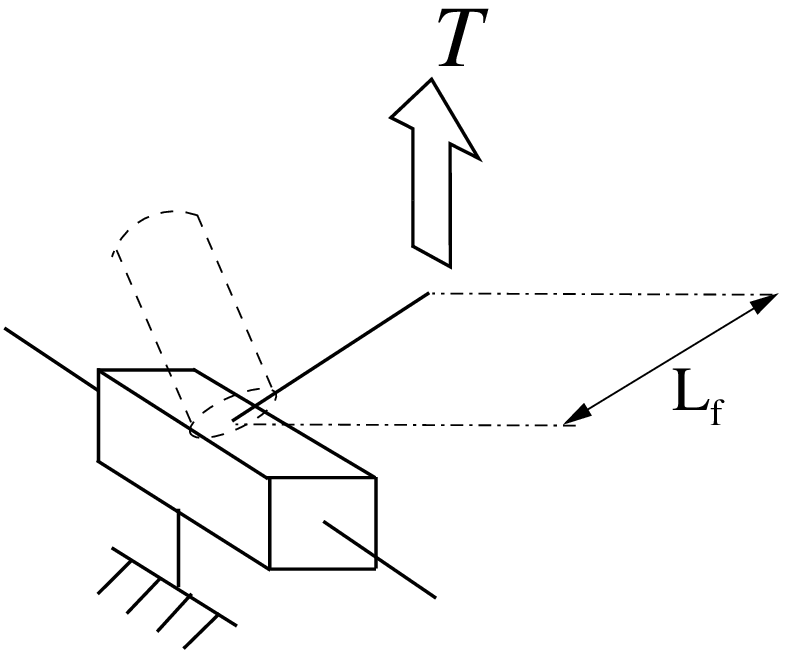}}}
\\  \hline

{$3EI_{f_1}/L_f$: foot bending due to force $F$ \newline {~}
\newline } &
{\scalebox{.5}{\includegraphics{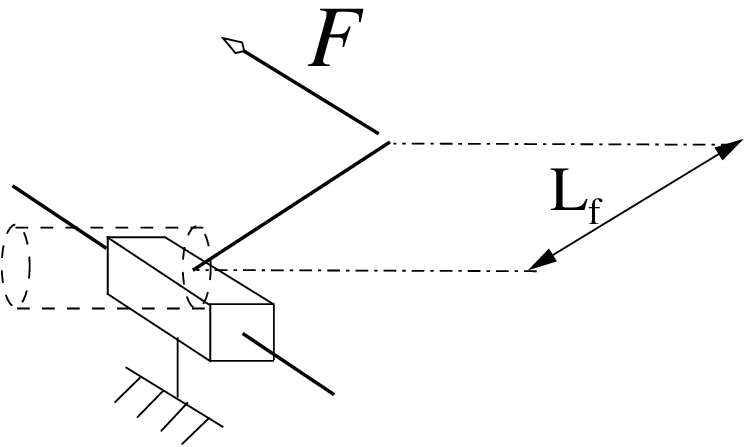}}}\\
\hline

{$ES_Bd^2/2L_B$: differential tension of parallelogram bars due to
torque $T$\newline {~} } &
{\scalebox{.4}{\includegraphics{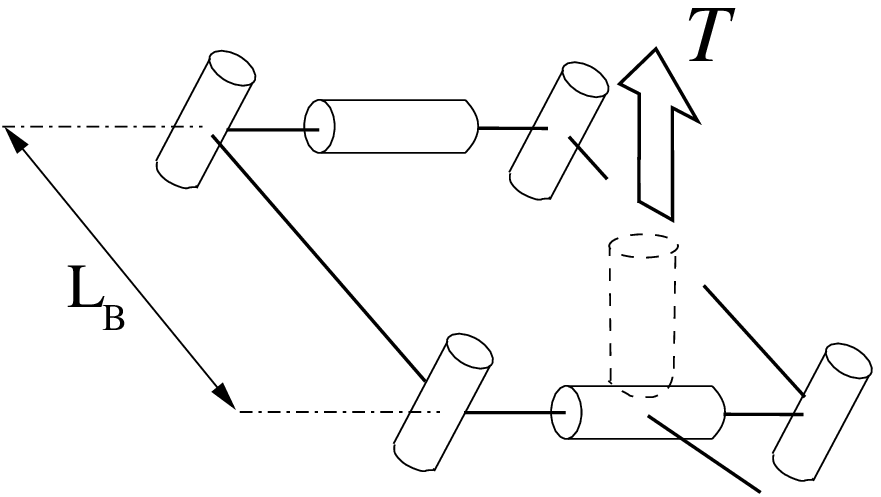}}}
\\

\hline

    \end{tabular}}
    \caption{Virtual compliant joints modeling}
    \label{tab_flex_first}
\end{center}
\end{table}

\section{Symbolic computing of the Orthoglide stiffness matrix}
\label{stiff_model}

\begin{figure}[ht!]
\begin{centering}
        \resizebox{!}{3.5cm}{\includegraphics{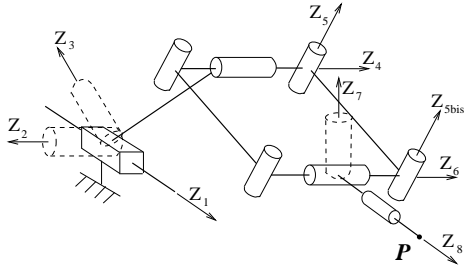}}
        \caption{Flexible leg}
        \label{jambe_ortho_flex}
\end{centering}
\end{figure}

The virtual compliant joints stiffnesses depend on the design
parameters. In this section, symbolic expressions of stiffness
matrix elements are symbolically computed. This part is based on a
stiffness model that was fully described in \cite{Gosselin02}.
Therefore, the description of the model will only be summarized
here. Fig. \ref{jambe_ortho_flex} represents the lumped model of a leg with flexible links.\\

The Jacobian matrix ${\bf{J}}_i$ of the $i$th leg of the
Orthoglide is obtained from the Denavit-Hartenberg parameters of
the $i$th leg with flexible links. This matrix maps all leg joint
rates (including the virtual compliant joints) into the
generalized velocity of the platform, i.e.,
\[ {\bf J}_i {\dot \theta}_i = {\bf t} \]
\noindent where \\
${\dot{\bf{\theta}}}_i^T = \left[\begin{array}{cccccccc}
{\dot{\bf{\theta}}}_{i_1} & {\dot{\bf{\theta}}}_{i_2} &
{\dot{\bf{\theta}}}_{i_3} & {\dot{\bf{\theta}}}_{i_4} &
{\dot{\bf{\theta}}}_{i_5} & {\dot{\bf{\theta}}}_{i_6} &
{\dot{\bf{\theta}}}_{i_7} & {\dot{\bf{\theta}}}_{i_8}
\end{array}\right]$ \\
is the vector containing the 8 actuated,
passive and compliant joint rates of leg i, $\bf t$ is the twist
of the platform. The $P_a$ joint parameterization imposes
${\dot{\bf{\theta}}}_{i_5}= - {\dot{\bf{\theta}}}_{i_{5bis}}$,
which makes ${\dot{\bf{\theta}}}_{i_5}$ and
${\dot{\bf{\theta}}}_{i_{5bis}}$
 dependent. ${\dot{\bf{\theta}}}_{i_5}$ is chosen to model
the circular translational motion, and finally ${\bf J}_i$ is
written as
$$
{\bf J}_i = \left[\begin{array}{cccc}
0 & {\bf e}_{i_2} & {\bf e}_{i_3} & {\bf e}_{i_4} \\
{\bf e}_{i_1} & {\bf e}_{i_2}\times{\bf r}_{i_2} & {\bf
e}_{i_3}\times{\bf r}_{i_3} & {\bf e}_{i_4}\times{\bf
r}_{i_4}\\
\end{array}
\right.
$$
$$
\left. \begin{array}{cccc}
 0 & {\bf e}_{i_6} & {\bf e}_{i_7} & {\bf e}_{i_8} \\
{\bf e}_{i_5}\times{\bf r}_{i_5}-{\bf e}_{i_{5bis}}\times{\bf
r}_{i_{5bis}} & {\bf e}_{i_6}\times{\bf r}_{i_6} & {\bf
e}_{i_7}\times{\bf r}_{i_7} & {\bf e}_{i_8}\times{\bf r}_{i_8} \\
                    \end{array}\right]
$$

\noindent  in which ${\bf e}_{i_j}$ is the unit vector along joint
j of leg i and ${\bf r}_{i_j}$ is the vector connecting joint j of
leg i to the platform reference point. Therefore the Jacobian
matrix of the Orthoglide can be written as:

\[
{\bf J} = \left[\begin{array}{ccc}
                 {\bf J}_1 & 0 & 0 \\
                  0 & {\bf J}_2 & 0  \\
                  0 & 0 & {\bf J}_3  \\
                  \end{array}\right]
\]

One  then has:

\begin{equation}
\label{eq_J}
{\bf J}\dot{\bf \theta} = {\bf R}{\bf t}
\end{equation}

\[
\mbox{with} \quad {\bf R}=\left[\begin{array}{c}
                    I_6 \\
                    I_6 \\
                    I_6
\end{array}\right]^T
\quad \mbox{and} \quad {\bf t} = \left\{\begin{array}{c} \Omega \\
{\bf V}
\end{array}\right\}
\]

\noindent $\dot{\bf{\theta}}$ being the vector of the 24 joint
rates, that is ${\dot{\bf{\theta}}} = \left[\begin{array}{ccc}
{\dot{\bf{\theta}}}_1^T & {\dot{\bf{\theta}}}_2^T &
{\dot{\bf{\theta}}}_3^T
\end{array}\right]$.  Unactuated joints are then eliminated
by writing the geometric conditions that constrain the two
independent closed-loop kinematic chains of the Orthoglide
kinematic structure :

\begin{equation}
\label{closchain1}
{\bf J}_1{\dot{\bf \theta}}_1 = {\bf
J}_2{\dot{\bf \theta}}_2
\end{equation}
\begin{equation}
\label{closchain2}
{\bf J}_1{\dot{\bf \theta}}_1 = {\bf
J}_3{\dot{\bf \theta}}_3
\end{equation}

\noindent From (\ref{closchain1}) and (\ref{closchain2})
 one can obtain
(see \cite{Gosselin02} for details):

\[ {\bf A}{\bf {\dot {\theta}'}}= {\bf B}{\bf \dot {\theta}''}\]

\noindent where $\bf \dot {\theta}'$ is the vector of joint rates
without passive joints and  $\bf \dot {\theta}''$ is the vector of
joint rates with only passive joints. Hence:

\[\dot {\theta}'' = {\bf B}^{-1}{\bf A} \dot {\theta}' \]

Then a  matrix $\bf{V}$ is obtained (see \cite{Gosselin02} for details) such that:

\begin{equation}
\label{eq_V}
\dot {\bf \theta}={\bf V} \dot {\bf \theta}'
\end{equation}

\noindent From (\ref{eq_J})
 and (\ref{eq_V}) one can obtain:

\begin{equation}
\label{eq_JV}
 {\bf J}{\bf V}\dot{\theta}' = {\bf R}{\bf t}
\end{equation}

\noindent As matrix ${\bf R}$ represents a system of 18 compatible
linear equations in 6 unknowns, one can use the least-square
solution to obtain an exact solution from (\ref{eq_JV}):

\[{\bf t} = ({\bf R}^T{\bf R})^{-1}{\bf R}^T{\bf J}{\bf V}\dot{\theta}'\]

Now let  ${\bf J}'$ be represented as:

\[{\bf J}' = ({\bf R}^T{\bf R})^{-1}{\bf R}^T{\bf J}{\bf V}\]

\noindent Then one has:

\begin{equation}
\label{eq_t}
{\bf t} = {\bf J}'\dot{\theta}'
\end{equation}

According to the principle of virtual work, one has:

\begin{equation}
\label{pvw}
{\bf \tau}^T {\dot \theta}' = {\bf w}^T{\bf t}
\end{equation}

\noindent where $\tau$ is the vector of forces and torques applied
at each actuated or virtual compliant joint and $\bf w$ is the
external wrench  applied at the end effector, point {\itshape
\textbf{P}}. Gravitational forces are neglected. By substuting
(\ref{eq_t}) in (\ref{pvw}), one can obtain:

\begin{equation}
\label{eq_tau}
\tau = {\bf J}'^T {\bf w}
\end{equation}

\noindent The forces and displacements of each actuated or virtual
compliant joint can be related by Hooke's law, that is  for the
whole structure one has:

\begin{equation}
\label{hooke}
{\bf \tau} =  {\bf K}_J \Delta \theta'
\end{equation}

\noindent with:

\[
{\bf K}_J   = \left[\begin{array}{ccc}
                 {\bf A} & 0 & 0 \\
                  0 & {\bf A} & 0  \\
                  0 & 0 & {\bf A}  \\
                  \end{array}\right]\\
\]

\noindent and:
\begin{equation}
\label{matA}
{\bf A} = \left[\begin{array}{cccc}
                  k_{act} & 0 & 0 & 0 \\
                  0 & \frac{3EI_{f_1}}{L_f} & 0 & 0  \\
                  0 & 0 & \frac{2EI_{f_2}}{L_f} & 0 \\
                  0 & 0 & 0 & \frac{ES_Bd^2}{L_B} \\
                  \end{array}\right]
\end{equation}

$\Delta \theta'$ only includes the actuated and virtual compliant
joints, that is by equating (\ref{eq_tau}) with (\ref{hooke}):

\[{\bf K}_J \Delta \theta'={\bf J}'^T{\bf w}\]

 \noindent hence:

\[ \Delta \theta'={\bf K}_J^{-1}{\bf J}'^T{\bf w}\]

\noindent Pre-multiplying both sides by ${\bf J}'$ one obtains:

\begin{equation}
\label{J_delta}
 {\bf J}'\Delta \theta'={\bf J}'{\bf K}_J^{-1}{\bf J}'^T{\bf w}
\end{equation}

\noindent Substituting (\ref{eq_t}) into (\ref{J_delta}), one
obtains:
\[  {\bf t}={\bf J}'{\bf K}_J^{-1}{\bf J}'^T{\bf w}\]
\noindent Finally the compliance matrix $\kappa$ is obtained as
follows:
\[
{\bf \kappa} ={\bf J}'{{\bf K}_J}^{-1}{\bf J}'^T
\]
In the Orthoglide case we obtain a simplified compliance matrix:

\begin{equation}
\label{stiffmat}
{\bf \kappa} = \left(\begin{array}{cccccc}
                  {\kappa}_{11} & 0 & 0 & 0 & {\kappa}_{15} & {\kappa}_{15}\\
                  0 & {\kappa}_{11} & 0 & {\kappa}_{24} & 0 & {\kappa}_{26}  \\
                  0 & 0 & {\kappa}_{11} & 0 & {\kappa}_{35} & 0  \\
                  0 & {\kappa}_{24} & 0 & {\kappa}_{44} & 0 & {\kappa}_{46} \\
                  {\kappa}_{15} & 0 & {\kappa}_{35} & 0 & {\kappa}_{55} & 0 \\
                  {\kappa}_{15} & {\kappa}_{26} & 0 & {\kappa}_{46} & 0 & {\kappa}_{66}
                  \end{array}\right)
\end{equation}

And the Cartesian stiffness matrix is:
\[
{\bf K}  = {{\bf J}'{{\bf K}_J}^{-1}{\bf J}'^T}^{-1}
\]

\noindent {\bf Remarks:}

From the expression of matrix ${\bf A}$ (see eq. \ref{matA}) we
deduce that all elements of stiffness matrix ${\bf K}$ can be
factored either in factors of $E$ or in factors of $k_{act}$. As
$k_{act}$ is assumed infinite compared to other stiffnesses, the
elements of ${\bf K}$ which are factors of $k_{act}$ must be
eliminated. Also obviously when $E$ increases, ${\bf K}$ elements
increase too, which is in accordance with intuition. Young's
modulus $E$ should then be eliminated from ${\bf K}$ elements
expressions because its influence needs no further study.

In eq. (\ref{stiffmat}) one should notice that
${\kappa}_{11}={\kappa}_{22}={\kappa}_{33}$, and
${\kappa}_{15}={\kappa}_{16}$. Also, we noticed that when X=Y=Z we
have ${\kappa}_{44}={\kappa}_{55}={\kappa}_{66}$.\\

\section{Parametric stiffness analysis}
\label{results}

In this part, we study the in uence of the geometric parameters on the elements of the Orthoglide stiffness matrix. For a realistic inspection of the symbolic expressions of these elements, it is necessary to replace some parameters by numerical values, once the symbolic expressions have been obtained. Numerical values of the design parameters come from the Orthoglide prototype \cite{SiteWeb}.

\subsection{Qualitative analysis at the isotropic configuration}

There is an isotropic configuration in the Orthoglide workspace [7]. This configuration gives a good insight of the Orthoglide general performances. It happens when the tool point $P$ is at the intersection of the three actuated joint axes (see Fig. 1). In this configuration, the stiffness matrix $\bf K$ is diagonal and the symbolic expressions of its elements are simplified:
\[
K = E/2
\left(
\begin{array}{cccccc}
K_a & 1 & 1 & 1 & 1 & 1 \\
1 & K_a & 1 & 1 & 1 & 1 \\
1 & 1 & K_a & 1 & 1 & 1 \\
1 & 1 & 1 & K_b & 1 & 1 \\
1 & 1 & 1 & 1 & K_b & 1 \\
1 & 1 & 1 & 1 & 1 & K_b \\
\end{array}
\right)
\]

$E=2K_a$ being the torsional stiffness with 
\begin{equation}
Ka =  \left( \frac{LB}{d^2S_B} + \frac{3Lf cos^2 \lambda }{h_f b^3_f}
\right)^{-1}
\end{equation}

and $E=2K_b$ being the translational stiffness with
\begin{equation}
Kb =  
\left(
\frac{0.5 h^3_f b_f}{L^3_f (1 -cos^2 \lambda} + \frac{E}{2k_{act}}  
\right)^{-1}
\end{equation}

\begin{itemize}
\item
From the inspection of $K_a$ we note that increasing the cross-section areas $SB$ and $b_f h_f$ increases the torsional stiffness, which is in accordance with intuition. Increasing the distance $d$ between the parallelogram bars increases the torsional stiffness. Also, it is no surprise that increasing $L_f$ or $L_B$ decreases the torsional stiffness. It is no surprise that $K_a$ and $K_b$ increase when $E$ increases, and that $K_b$ increases when $k_{act}$ increases, while $Ka$ does not depend on $k_{act}$, the prismatic actuated joint stiffness. $K_a$ and $K_b$ do not depend on e at the isotropic configuration.
\item From the inspection of $K_b$ we note that increasing the cross-section area $b_fh_f$ increases the translational stiffness, as increasing the foot length $L_f$ does. We also note that if $\lambda$ (angle between the foot and the actuated joint axis) decreases toward zero, then the translational stiffness tends toward infinite. This is in accordance with intuition since at the isotropic configuration, if $\lambda = 0$ then the virtual compliant joint ``foot bending due to force $F$" (see Tab. 2) is not requested because the applied force axis crosses the virtual compliant joint axis. 

Furthermore, if $\lambda = 0$, the cantilever disappears. As this cantilever is aimed at preventing each parallelogram from colliding with the corresponding linear motion guide, from a designer point of view the conclusion could be: ``$\lambda$ should be lowered to increase the translational stiffness, but there should remain enough cantilever for the workspace volume to be large enough". 
\end{itemize}
\subsection{Quantitative analysis at the isotropic configuration}
\begin{figure}[hbt]
    \begin{center}
    \begin{tabular}{cc}
       \begin{minipage}[t]{80 mm}
           {\scalebox{.4}{\includegraphics{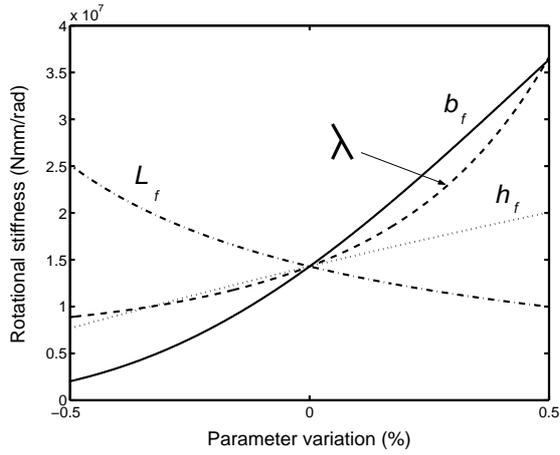}}} \\
(a) Influence of $L_f$ , $b_f$ , $h_f$ and $\lambda$ 
       \end{minipage} &
       \begin{minipage}[t]{80 mm}
           {\scalebox{.4}{\includegraphics{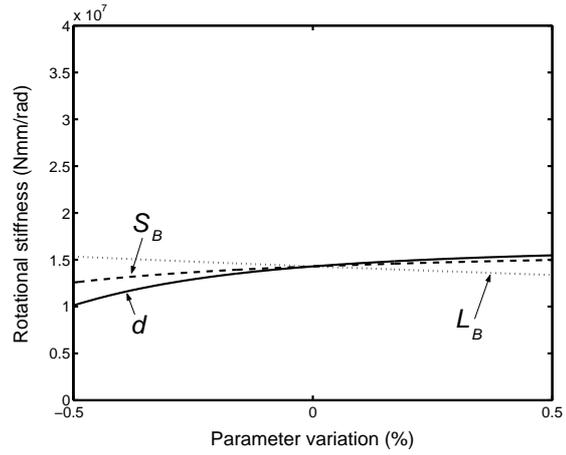}}} \\
(b) Influence of $d$, $L_B$ and $S_B$
       \end{minipage}
      \end{tabular}
    \end{center}
    \caption{Quantitative influence of the geometric parameters on the rotational stiffness $E=2K_a$}
    \label{figure06}
\end{figure}

\begin{figure}[hbt]
    \begin{center}
           {\scalebox{.4}{\includegraphics{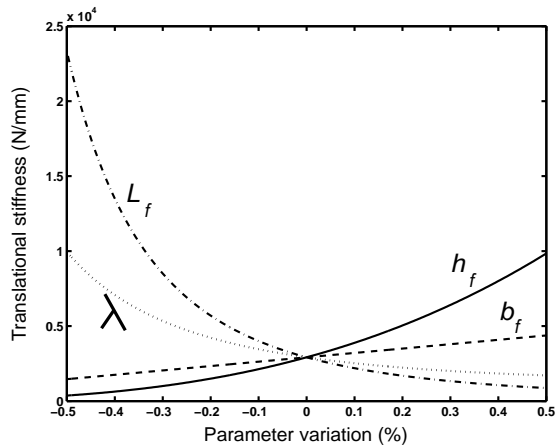}}} 
    \end{center}
    \caption{Quantitative influence of the geometric parameters $h_f$ , $b_f$ , $L_f$ and $d$ on the translational stiffness $E=2K_b$}
    \label{figure07}
\end{figure}

The qualitative analysis conducted above provides  an interesting insight of the influence of the geometrical parameters on rotational and translational stiffnesses. Quantitative information is now obtained by studying the influence on $K_a$ and $K_b$ of a $+/- 50\%$ variation of the numerical value of each design parameter. The results are shown on Fig.\ref{figure06} and on Fig.\ref{figure07}. From Fig. 6 analysis, we can go further than the qualitative analysis and see that $d$, $S_B$ and $L_B$ have very little influence on the rotational stiffness while $L_f$ , $h_f$ $b_f$ and $\lambda$ have much more influence. From Fig. 7 we note, as deduced from the qualitative analysis, that decreasing $L_f$ or $\lambda$ increases the translational stiffness, and that increasing hf can improve the translational stiffness. Increasing hf does not reduce the workspace volume like reducing $L_f$ does. However increasing hf increases the foot weight, which could be prejudicial to the dynamic performances. The proof of this forecast needs further investigations. Quantitative analysis provides another point of view and helps in a better understanding of the influence of the geometric design parameters on the machine stiffness.
\section{Conclusions}
A parametric stiffness analysis of a 3-axis PKM prototype, the Orthoglide, was conducted. First, a compliant model of the Orthoglide was obtained, then an existing stiffness analysis method for parallel manipulators was applied to the Orthoglide. The stiffness matrix elements were computed symbolically. In the isotropic configuration, the influence of the geometric design parameters on the rotational and translational stiffnesses was studied through qualitative and quantitative analyses. These provided relevant information for stiffness-oriented design or optimization of the Orthoglide.

This example of a parametric stiffness analysis shows that simple symbolic expressions carefully built and interpreted provide broad information on parallel manipulators stiffness features. Comparison of the stiffness model with a finite elements model and with stiffness tests on the prototype are currently being conducted.

\end{document}